\documentclass[letterpaper, 10 pt, conference]{ieeeconf}

\IEEEoverridecommandlockouts                          
\usepackage{amsmath} % assumes amsmath package installed
\usepackage{amssymb}  % assumes amsmath package installed
\usepackage{algorithm}
\usepackage{algpseudocode}
\usepackage{graphicx}
\usepackage{physics,bm,bbm}

\usepackage{multirow}
\usepackage{cuted}
\usepackage{cite}

\usepackage{booktabs}
\usepackage{colortbl}
\usepackage{multicol}
\usepackage{xcolor}
\usepackage{makecell}
\usepackage{subcaption}
\usepackage{enumitem}
\newcommand{\hide}[1]{}

\newcommand\blfootnote[1]{%
  \begingroup
  \renewcommand\thefootnote{}\footnote{#1}%
  \addtocounter{footnote}{-1}%
  \endgroup
}

\definecolor{es-blue}{rgb}{0,0.4,0.8}

\definecolor{lightgray}{gray}{0.9}

\newcommand{\upper}{\text{upper}}
\newcommand{\lowerr}{\text{lower}}

\newcommand{\method}{\textbf{PMP}}
\newcommand{\methodn}{\text{PMP}}
\newcommand{\motiondata}{\mathcal{M}}

\definecolor{cvprblue}{rgb}{0.21,0.49,0.74}
\usepackage[pagebackref,breaklinks,colorlinks=true,citecolor=cvprblue, urlcolor=blue]{hyperref}

\title{\LARGE \bf
Mobile-TeleVision: Predictive Motion Priors for \\  Humanoid Whole-Body Control
}

\author{
    Chenhao Lu$^{*1}$, Xuxin Cheng$^{*1}$, Jialong Li$^{*1}$, Shiqi Yang$^{1}$, Mazeyu Ji$^{1}$, Chengjing Yuan$^{1}$ \\ Ge Yang$^{2}$, Sha Yi$^{1}$, Xiaolong Wang$^{1}$ \\ [0.1in]
    \textsuperscript{1}UC San Diego, \textsuperscript{2}MIT\\ [0.1in]
    \url{https://mobile-tv.github.io/}
    \vspace{-0.2in}
}% <-this % stops a space

\begin{document}
\twocolumn[{%
\renewcommand\twocolumn[1][]{#1}%
\maketitle
\begin{center}
    % \vspace{-0.1in}
    \centering
    \captionsetup{type=figure}
    \includegraphics[width=\linewidth]{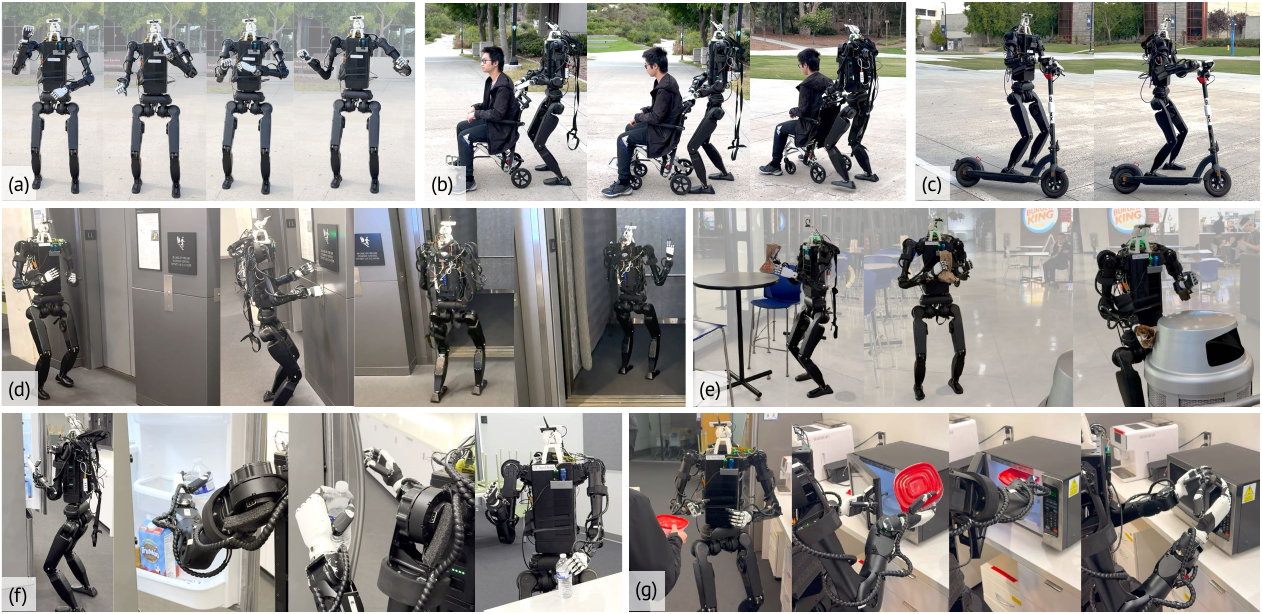}
    \caption{Humanoid robot doing whole-body tasks that require both precise manipulation and robust locomotion. The robot can a) dance with its arms while the lower body maintains balance, b) push a wheelchair with a human, c) walk while carrying a scooter, d) use an elevator, e) collect and throw away trash, f) open the fridge, take and deliver a bottle of water, and g) put a food storage container into the microwave. }
    \label{fig:teaser}
    \vspace{-0.1in}
\end{center}
}]
{\blfootnote{{\authorrefmark{1} Equal contribution. }}}
% \blfootnote{
% \\
% \textsuperscript{$1$} University of California San Diego \\
% \textsuperscript{$2$} Tsinghua University \\
% \textsuperscript{$\dag$} Work done during internship at UC San Diego.\\
% \textsuperscript{*} Equal contributions. \\
% }

\thispagestyle{empty}
\pagestyle{empty}

%%%%%%%%%%%%%%%%%%%%%%%%%%%%%%%%%%%%%%%%%%%%%%%%%%%%%%%%%%%%%%%%%%%%%%%%%%%%%%%%
\begin{abstract}

Humanoid robots require both robust lower-body locomotion and precise upper-body manipulation.
% to conduct diverse tasks. 
While recent Reinforcement Learning (RL) approaches provide whole-body loco-manipulation policies, they lack precise manipulation with high DoF arms. In this paper, we propose decoupling upper-body control from locomotion, using inverse kinematics (IK) and motion retargeting for precise manipulation, while RL focuses on robust lower-body locomotion. 
We introduce \textit{PMP (Predictive Motion Priors)}, trained with Conditional Variational Autoencoder (CVAE) to effectively represent upper-body motions. The locomotion policy is trained conditioned on this upper-body motion representation, ensuring that the system remains robust with both manipulation and locomotion. We show that CVAE features are crucial for stability and robustness, and significantly outperforms RL-based whole-body control in precise manipulation. With precise upper-body motion and robust lower-body locomotion control, operators can remotely control the humanoid to walk around and explore different environments, while performing diverse manipulation tasks.
\end{abstract}
%%%%%%%%%%%%%%%%%%%%%%%%%%%%%%%%%%%%%%%%%%%%%%%%%%%%%%%%%%%%%%%%%%%%%%%%%%%%%%%%
\section{Introduction}
\label{sec:intro}
Our ability to precisely grasp objects while being on the move is essential to our daily lives. Looking at humanoid control, what we demand of its arms is quite different from those of the legs: We want the arms to be versatile enough to apply variable forces, to hold objects at a fixed position and to move rapidly, such as when catching a ball. Meanwhile, its legs must also maintain balance. It should be able to walk while carrying a box, or to run while it dribbles. Current state-of-the-art humanoid whole-body control is unable to do this. The key missing piece is a way to integrate expressive and load-bearing movements of the upper body with control of the legs that fully captures common ranges of motion.

Recent advances in whole-body humanoid control using Reinforcement Learning (RL) have demonstrated the ability of robots to robustly execute a wide range of motions \cite{fu2024humanplus, cheng2024exbody, he2024omnih2o} from large-scale human motion dataset \cite{AMASS:2019} or teleoperation \cite{cheng2024tv, park2024avp}. RL commonly exposes the robot in simulation to a variety of target motions and failure modes. Such setups are able to produce performant policies that can directly deploy in the real world, but at the expense of expressiveness, due to Reinforcement Learning's tendency to seek and overfit to particular modes.
RL policy is also not well-suited for high DoF positional and orientation control \cite{liu2024visual, fu2022learning} and can output unpredictable actions for arms, especially in dynamic walking scenarios. It brings additional complexity to training the policy as well. 
These experimental results share a consensus with our observations that different mechanisms are required to control arms and legs. 

The upper body's movements can be efficiently computed using inverse kinematics (IK), without needing to address complex balance challenges. Based on this, we propose removing the upper-body tracking during RL training, leaving the RL controller responsible solely for robust walking based on velocity commands. This approach reduces exploration and training costs while ensuring high precision for the upper body using direct joint position control. 

While decoupling upper and lower bodies brings benefits to the manipulation precision, prior studies~\cite{cheng2024exbody, fu2022learning,he2024omnih2o} also show that fully decoupling the control of the upper and lower body may lead to instability or loss of balance, as it fails to account for the interaction between them. The goal of this work is to develop an integrated system that combines the precision and expressiveness of classical motion planning for the upper body, with the real-world robustness of locomotion control trained via deep reinforcement learning, in a single, whole-body control system. We do so by treating this as a representation learning problem, that starts with learning \textit{Predictive Motion Priors (PMP)} of the upper body movement using a conditional variational autoencoder (CVAE) to encode upper-body motions as the observation of the policy. Specifically, we first train a CVAE on a motion dataset. Conditioned on past motion frames, the decoder of the CVAE predicts future motion frames with a randomly sampled latent vector. Next, we train an RL locomotion policy that observes the latent vector as an observation. The output of the RL policy only involves motors for the lower body and the upper body motor angles can be set in a general way such as IK or motion retargeting, resulting in higher precision.

This decoupled structure is particularly well-suited for teleoperation systems designed for loco-manipulation tasks, where the operator can control the robot's mobility using simple velocity commands while simultaneously performing precise arm and hand operations. By leveraging inverse kinematics (IK) and hand re-targeting, the teleoperator can efficiently perform complex manipulation tasks without the need to focus on balancing or lower-body coordination. As shown in Figure~\ref{fig:teaser}, this enables us to perform fine, dynamic, and load-bearing manipulation tasks more efficiently, even when the robot is on the move. We compare our method and previous approaches in detail in Table \ref{tab:comparisons}. 

To test the generalizability of our method, we experiment with both the Fourier GR1 and the Unitree H1 robots in simulation, and evaluate the Unitree controller in the real world. We test its robustness by performing large motions in a teleoperation setup that involves the robot bearing various loads (Figure~\ref{fig:teaser}). Our result showcases PMP's ability to realize accurate upper-body control for a pair of high-DoF arms while remaining robust, a new capability previously unseen by prior methods.

\begin{table}[t]
\resizebox{\linewidth}{!}{%
\begin{tabular}{c|c|c|c|c}
\toprule
     Metrics & \makecell{\method \\ (Ours)} & \makecell{ExBody\\ \cite{cheng2024exbody}} & \makecell{OmniH2O \\ \cite{he2024omnih2o}}  & \makecell{HumanPlus \\ \cite{fu2024humanplus}} \\
    \midrule
    Arm DoFs & 7 & 4 & 5 & 5 \\
    PrManip &\checkmark &$\times$ & $\times$ & $\times$\\
    Robust Walking &\checkmark &\checkmark & \checkmark & $\times$\\
\bottomrule
\end{tabular}
}
    \caption{Comparisons with whole-body control methods. PrManip means whether arm motions are directly controlled and do not go through RL mapping. Robust walking means whether the work shows at least 10 seconds of robot walking. }
    \vspace{-10pt}
    \label{tab:comparisons}
    % \vspace{-0.1in}
\end{table}

\begin{figure*}[htbp]
\centering
\includegraphics[width=\linewidth]{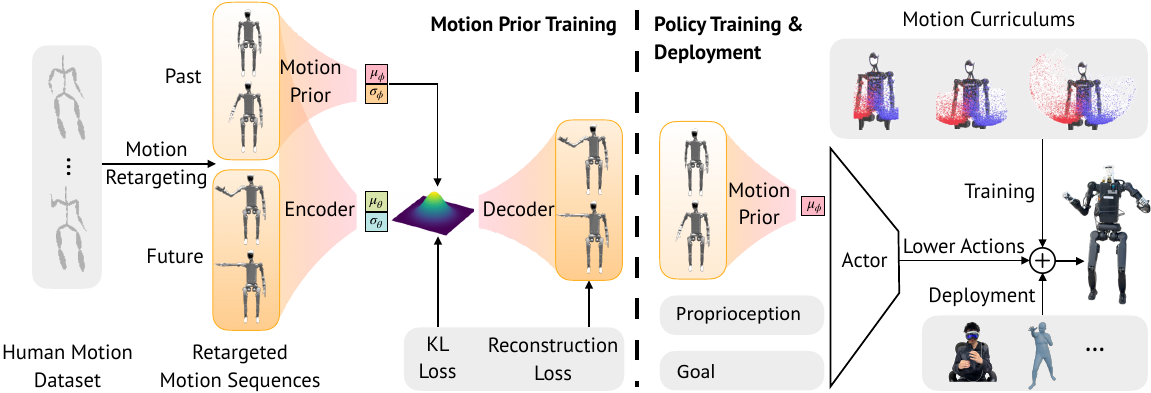}
\caption{
The training pipeline consists of three stages: (a) preprocessing of the motion dataset by mapping local rotation, (b) training a CVAE to capture prior knowledge of upper body human motion, and (c) RL training where the upper body is controlled using sampled target joint positions and the lower body is trained using prior motion representations. 
}
\label{fig:method}
\vspace{-0.1in}
\end{figure*}

\section{Problem Formulation}\label{sec:problem_form}

We focus on providing a lower-body controller that tracks a given body velocity while maintaining stability when the upper body performs dynamic actions. 

We consider this problem as a goal-conditioned motor policy $\pi:\mathcal{G} \times \mathcal{S} \rightarrow \mathcal{A}$. 
The goal space $\mathcal{G}$ includes the target behavior of both the root and the upper body. $\mathcal{S}$ consists of proprioception and auxiliary information obtained from observations. The auxiliary information in our work is the motion prior, which we will go into detail in the next section. $\mathcal{A}$ is the action space, which contains the PD torque control input for each joint.

\textit{a) Goal Space for Locomotion Control and Upper Body Motion }: The goal space $\mathcal G$ contains two parts: the locomotion goal $\mathcal G^m=\expval{\vb v,rpy,h}$ for the lower body, and the upper body motion goal $\mathcal G^u=\expval{\vb q^\upper}$. The base locomotion is to track the given linear velocity $\vb v\in\mathbb R^3$, orientation described by Euler angles roll $r$, pitch $p$, and yaw $y$, and the base height $h$. $\vb q^\upper$ is the target position of upper body joints including the waist, shoulders, elbows, wrists, and hands. For the target platforms of this paper, H1 (with dexterous hands) and GR1 contain $27$ and $19$ upper-body joints, respectively.

\textit{b) Decoupled control policy}: We decouple the control policy by $\pi=[\pi^\upper,\pi^\lowerr]$. $\pi^\upper$ is the open loop control, directly outputting the upper body motion goal $\vb q^\upper$. $\pi^\lowerr$ is trained by RL. It observes proprioception $\vb s_t=[\vb q_t,\dot{\vb q}_t,\bm{\omega}_t,\vb{g}_t,\vb{a}_{t-1}]$ and the locomotion command where $\vb{q}_t,\dot{\vb{q}}_t$ is current joint positions and velocities, $\bm{\omega}_t \in \mathbb R^3$ is current base angular velocity, $\vb{g}_t$ is current projected gravity, $\vb{a}_{t-1} \in \mathbb{R}^{12}$ is the previous action. Both GR1 and H1 have 12 joints in the lower body. Our method takes auxiliary information as input including gait periodic signals and motion prior. 

\section{Method}
\label{sec:method}
As illustrated in Figure \ref{fig:method}, our training pipeline consists of three stages. The first stage involves preprocessing a given human motion dataset, where we apply a local rotation mapping to convert the local rotation matrices into joint angles on the robots. We follow the data filtering and retargeting process described in ExBody~\cite{cheng2024exbody} to get a retargeted motion dataset $\motiondata$. In the second stage, we train the model to autoregressively generate future motions based on the retargeted dataset. The generated motion sequence is then used to extract motion priors, which we introduce in detail in section~\ref{sec:motion_prior}. In the third stage, we directly control the upper body using sampled target joint positions in a curriculum schedule. This is then combined with the lower body control for whole-body movements.

\subsection{Motion Prior Training}
\label{sec:motion_prior}

Since our control policy decouples the upper and lower body, the robustness of the lower body control becomes critical when the upper body controller has separate policies. The key to addressing this challenge lies in informing the lower body control with prior knowledge of upper body movements. 

Specifically, we estimate the future motions of the upper body from previous motions based on the retargeted dataset, and train a representation to be incorporated in the state space of the lower body control. We define this representation as \textit{Predictive Motion Priors} (PMP). 

Conditional variational autoencoder (CVAE) has been shown as an effective motion representation~\cite{rempe2021humor, luo2023universal}.
A CVAE architecture contains a prior model $R$, an encoder $E$, and a decoder $D$. Given an upper-body motion sequence within a time frame of $t-W$ to $t+W$, where $W$ is a time window, the prior model $R$ calculates a prior distribution for a latent vector $z_t$. It allows us to generate the essential future upper-body motion from $z_t\in\mathbb R^H$. $z_t$ is the desired motion prior for our policy.

Formally, we model the encoder and prior distribution as diagonal Gaussian: \begin{align}
    R(\cdot|M_t^0)&=\mathcal N\qty(\cdot|\mu_\phi(M_t^0),\sigma_\phi(M_t^0))\\
    E(\cdot|M_t^0,M_t^1)&=\mathcal N\qty(\cdot|\mu_\theta(M_t^0,M_t^1),\sigma_\theta(M_t^0,M_t^1)),
\end{align}
where $M_t^0=\vb q_{t-W:t-1}^\upper$ and $M_t^1=\vb q_{t:t+W-1}^\upper$ are two consecutive motion sequence.

During training, $(M_t^0,M_t^1)$ are sampled from the dataset $\motiondata$. The objective in the evidence lower bound is as follows: \begin{align*}
\log P(M_t^1 | M_t^0) &\ge \mathbb{E}_{z_t \sim E}\left[D(M_t^1 | z_t, M_t^0)\right] \\
&\quad - \operatorname{KL}\left(E(z_t | M_t^0, M_t^1) \parallel R(z_t | M_t^0)\right).
\end{align*}

In the specific implementation, the structures of $R$, $E$, and $D$ are three-layer MLPs, and the size $H$ of $z_t$ is $64$. 
The decoder $D$ is responsible for reconstructing the future motion sequence $M_t^1$ from the latent vector $z_t$. The decoder learns to map $z_t$ back into the space of motion sequences, aiming to generate a future motion sequence $M_t^1$ that closely resembles the original future motion sequence.
The window length $W$ corresponds to a real time of $1s$. Due to different control frequencies, $W$ is $50$ for H1 and $100$ for GR1.
In the next stage, we use the prior model $R$ to calculate motion prior $z_t$, which is as state input for the control policy.

\subsection{Decoupling Policy Training}

We use PPO~\cite{schulman2017proximal} to train the lower-body policy. During training, the environment will randomly sample a motion sequence $M_t=\vb q^\upper_{1:T}$ from the retargeted motion dataset $\motiondata$ at the beginning of a new episode. At step $t$ , $\vb q_{1:T}^\upper$ will be directly set as the PD target of the humanoid's upper body.

To overcome the exploration burden for RL with hard motions, we introduce an adaptive difficulty factor $\alpha_i$ for each motion $i$, adjusting the amplitude of the upper body motions in a curriculum schedule.

Given a sequence of upper-body motion $i$, denote the target joint position to be $\vb q^{\upper}_\text{target}$. For the PD controller of the joint position during the curriculum, we follow
\begin{equation}
\vb q^\upper_\text{curriculum}=\vb q_0 + \alpha_i(\vb q^{\upper}_\text{target} - \vb q_0)
\end{equation}
where $\vb q_0$ is the default joint position for the robot.

The difficulty factor $\alpha_i$ changes when the environment is reset with the following rule:
\begin{equation}
\alpha_i \leftarrow \begin{cases}
    \alpha_i + 0.05 & \text{survival time $\geq$ $90\%$ total time}\\
    \alpha_i - 0.01 & \text{otherwise}
\end{cases}
\end{equation}

For reward design, we follow ExBody~\cite{cheng2024exbody} with additional explicit gait periodicity rewards, as seen in previous work~\cite{zhang2024whole,margolis2023walk}.

\begin{table*}[t]
    \centering
    \resizebox{\textwidth}{!}{
\begin{tabular}{lrrrrrrrrr}
\toprule
Method & $\mathrm{E}_\text{acc}^\text{upper}$ $\downarrow$ & $\mathrm{E}_\text{action}^\text{upper}$ $\downarrow$ & $\mathrm{E}_\text{kpe}^\text{upper}$ $\downarrow$ & $\mathrm{E}_\text{jpe}^\text{upper}$ $\downarrow$ & $\mathrm{E}_\text{vel}$ $\downarrow$ & $\mathrm{E}_\text{ang}$ $\downarrow$ & $\mathrm{E}_\text{acc}^\text{lower}$ $\downarrow$ & $\mathrm{E}_\text{action}^\text{lower}$ $\downarrow$ & $\mathrm{E}_\text{g}$ $\downarrow$ \\
\cmidrule(r){1-1} \cmidrule(r){2-3} \cmidrule(r){4-5} \cmidrule(r){6-7} \cmidrule(r){8-10}
\rowcolor{lightgray}
\multicolumn{10}{l}{\textbf{(a) Simulation results for H1}} \\
\cmidrule(r){1-1} \cmidrule(r){2-3} \cmidrule(r){4-5} \cmidrule(r){6-7} \cmidrule(r){8-10}
Ours (\methodn) & \textbf{5.07} & \textbf{0.00280} & 0.02161 & \textbf{0.02355} & 0.01296 & \textbf{0.01947} & 13.28 & 0.01651 & \textbf{1.301} \\
Exbody & 11.93 & 0.01595 & 0.02303 & 0.03125 & 0.00784 & 0.02105 & 14.69 & 0.01651 & 1.367 \\
Exbody (Whole) & 9.97 & 0.01576 & 0.02474 & 0.03788 & \textbf{0.00782} & 0.02051 & 14.76 & \textbf{0.01553} & 1.454 \\
\cmidrule(r){1-1} \cmidrule(r){2-3} \cmidrule(r){4-5} \cmidrule(r){6-7} \cmidrule(r){8-10}
Ours w/o motion prior & 5.24 & 0.00282 & \textbf{0.02146} & 0.02425 & 0.01349 & 0.02081 & \textbf{13.27} & 0.01623 & 1.585 \\
Ours w/o curriculum & 5.39 & 0.00289 & 0.02150 & 0.02440 & 0.01481 & 0.02057 & 13.54 & 0.01645 & 1.597 \\
\cmidrule(r){1-1} \cmidrule(r){2-3} \cmidrule(r){4-5} \cmidrule(r){6-7} \cmidrule(r){8-10}
\rowcolor{lightgray}
\multicolumn{10}{l}{\textbf{(b) Simulation results for GR1}} \\
\cmidrule(r){1-1} \cmidrule(r){2-3} \cmidrule(r){4-5} \cmidrule(r){6-7} \cmidrule(r){8-10}
Ours (\methodn)& \textbf{3.08} & \textbf{0.00054} & \textbf{0.02165} & \textbf{0.00730} & 0.00409 & \textbf{0.00886} & \textbf{7.83} & 0.00784 & \textbf{0.420} \\
Exbody (Reimplementation) & 7.47 & 0.00200 & 0.02282 & 0.01006 & 0.00978 & 0.01448 & 9.74 & \textbf{0.00425} & 0.835 \\
\cmidrule(r){1-1} \cmidrule(r){2-3} \cmidrule(r){4-5} \cmidrule(r){6-7} \cmidrule(r){8-10}
Ours w/o motion prior & 3.11 & \textbf{0.00054} & 0.02166 & 0.00730 & \textbf{0.00393} & 0.00899 & 8.07 & 0.00773 & 0.442 \\
Ours w/o curriculum & 3.38 & 0.00055 & 0.02166 & 0.00737 & 0.00424 & 0.01105 & 9.24 & 0.00916 & 0.546 \\
\bottomrule
\end{tabular}
    }
    \caption{Comparisons with baselines. We sample $5$ trajectories for each motion from the dataset in simulation and report their mean episode metrics. The locomotion commands are from the base state of the corresponding motion.}
    \label{tab:result-sim}
\end{table*}
\section{Experiments}
\label{sec:exp}
We use the Unitree H1 robot for our real-world experiments and an additional Fourier GR-1 robot for simulation experiments. The details of the H1 robot are shown in Figure \ref{fig:hardware} right.
We aim to answer the following questions by both simulation and real-world experiments: \begin{enumerate}
    \item How well does \method~perform compared to those methods in learning-based upper-body control?
    \item  How does the tracking precision and the stability of \method~ perform when a disturbance occurs?
    \item How does \method ~perform in real-world scenarios requiring both robust locomotion and precise manipulation?
\end{enumerate}

\subsection{Simulation Results}
\begin{figure*}[htbp]
    \centering
    \includegraphics[width=\linewidth]{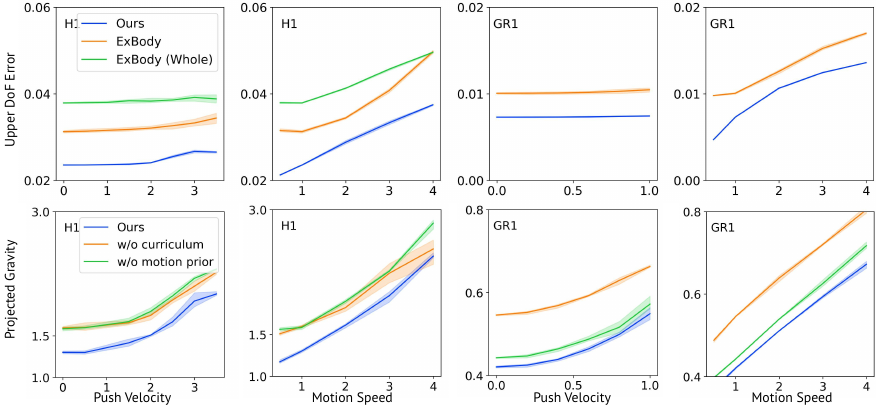}
    \caption{Evaluation for precision (Top) and stability (Bottom) under disturbance (lower is better for all figures). We sample $5$ trajectories for each motion from the motion dataset in simulation and report their mean episode metrics for each type of disturbance. The locomotion commands are from the base state of the corresponding motion. The robot is pushed every 5 seconds by a sudden increase in velocity to a value of $\texttt{push\_vel}$. Motion speed is changed by multiplying the frame rate of retargeted motion by the corresponding factor. PMP shows lower values, which correspond to better performance.}
    \label{fig:robust-sim}
    \vspace{-0.1in}
\end{figure*}

To answer \textbf{Q1} in simulation, the performance of \method~is evaluated by comparing it with the following baseline: \begin{itemize}
    \item \textbf{Exbody}: This baseline's upper body control policy is RL-trained with tracking rewards for upper body target keypoints and joint positions. The code for H1 is sourced from the Exbody's~\cite{cheng2024exbody} codebase. The code for GR1 is sourced from our own reimplementation.
    \item \textbf{Exbody (Whole)}: This baseline's upper body control policy is RL-trained with tracking rewards for whole-body target keypoints and joint positions. The H1 code is sourced from Exbody's codebase, and similar methods include HumanoidPlus~\cite{fu2024humanplus} and OmniH2O~\cite{he2024omnih2o} in terms of reward settings and RL training.
    \item \textbf{\method \space without motion prior}: This baseline uses decoupled policy but without motion prior. 
    \item \textbf{\method \space without motion curriculum}: This is an ablation study on motion curriculum. In this baseline, the $\alpha$ in the motion curriculum is always set to 1.
\end{itemize}

Our metrics are as follows: \begin{enumerate}
    \item \textit{Precision}: upper joint position error $E_\text{jpe}^\upper$, upper key point position error $E_\text{kpe}^\upper$.
    \item \textit{Smoothness}: upper joint acceleration $E^\upper_\text{acc}$, upper action difference $E^\upper_\text{action}$ (difference between consecutive actions, measuring motion smoothness).
    \item \textit{Locomotion}: command linear velocity error $E_\text{vel}$, command angular velocity error $E_\text{ang}$.
    \item \textit{Stability}: lower joint acceleration $E^\lowerr_\text{acc}$, lower action difference $E^\lowerr_\text{action}$ (difference between consecutive actions), projected gravity $E_\text{g}$ (gravity projection on the robot’s body $z$-axis, indicating balance).
\end{enumerate}

Results are shown in Table~\ref{tab:result-sim}. We observe that the control policy employing the decoupling method proposed in this paper demonstrates significantly higher \textit{precision} and \textit{smoothness} in tracking upper-body motion targets compared to the use of a learnable policy for the upper body. For \textit{locomotion} and \textit{stability}, results from the H1 reveal that the decoupling strategy leads to a decline in performance metrics for motion command tracking and lower-body stability, due to the common trade-offs in loco-manipulation. However, \methodn, by leveraging motion prior and motion curriculum, improves lower-body stability, achieving performance metrics comparable to the best baseline.

To answer \textbf{Q2}, we conduct robustness tests under perturbations such as pushing the robot and varying the playback speed of upper-body motions. We assess precision by the average upper DoF error $E_\text{jpe}^\upper$ and stability by the average projected gravity $E_\text{g}$. The results are presented in Figure~\ref{fig:robust-sim}. As disturbances increase, the precision comparison shows that PMP consistently exhibits superior tracking accuracy compared to methods like ExBody by about $40\%$, showcasing the advantage of decoupled upper-body control. Regarding stability, we observe that the introduction of motion prior and motion curriculum significantly improves the policy’s robustness by about $25\%$ in terms of projected gravity. Overall, our results demonstrate that \textit{PMP} achieves precise upper-body control while maintaining robustness.

\subsection{Teleoperation Setup}

\begin{table}[t]
    \hspace*{-\leftmargin}
    \centering
    \begin{subtable}{.45\textwidth}
    \centering
    \begin{tabular}{ccc}
    \toprule
     Metrics & Ours & Ours w/o motion prior \\
    \midrule
    Max (abs) pitch vel (rad/s) & \textbf{0.253} & 0.296 \\
    Pitch deviation range (rad) & (-0.152, \textbf{0.035}) & (\textbf{-0.138}, 0.050)\\
    Recovery time (s) & \textbf{0.59} & 0.66 \\
    \bottomrule
    \end{tabular}
    \caption{Robustness testing results - walking}
    \label{tab:robustness-walking}
    \end{subtable}
    \hspace*{-\leftmargin}
    \begin{subtable}{.45\textwidth}
    \centering
    \begin{tabular}{ccc}
    \toprule
     Metrics & Ours & Ours w/o motion prior \\
    \midrule
    Max (abs) pitch vel (rad/s) & \textbf{0.241} & 0.331 \\
    Pitch deviation range (rad) & \textbf{(-0.150, 0.029)} & (-0.161, 0.051)\\
    Recovery time (s) & \textbf{4.46} & 5.38\\
    \bottomrule
    \end{tabular}
    \caption{Robustness testing results - standing}
    \label{tab:robustness-standing}
    \end{subtable}
    \caption{Robustness testing results; each number is averaged over ten pushes. Recovery time represents the time elapsed before the robot can steadily walk or stand in a stationary location again.}
    \label{tab:robustness}
    \vspace{-0.1in}
\end{table}

\begin{table}[t]
    \centering
    \centering
    \begin{tabular}{p{0.4in}cccccc}
    \toprule
     Task & x(m) & y(m) & z(m) & roll(rad) & pitch(rad) & yaw(rad)\\
    \midrule
    Sorting & 0.011 & -0.032 & 0.008 & 0.002 & 0.040 & 0.016\\
    Insertion & 0.027 & -0.018 & 0.006 & -0.021 & -0.009 & 0.036\\
    Folding & 0.044 & -0.018 & 0.002 & -0.013 & -0.010 & 0.171 \\
    Unloading & 0.027 & -0.028 & 0.008 & -0.045 & 0.040 & 0.093\\
    \bottomrule
    \end{tabular}
    \caption{Arm tracking errors. For each task, the errors are averaged over twenty trajectories, the equivalency of roughly 30 minutes of arm movement under 50 Hz control frequency.}
    \vspace{-0.1in}
    \label{tab:precision}
\end{table}

To test the performance of our method in the real world, we set up a real-world teleoperation system with the Unitree H1 robot. The teleoperation of the active neck and upper body follows the pipeline described in \cite{cheng2024tv}. The operator uses the Apple Vision Pro to track hand and head movements, which are then translated into the robot's joint angles through inverse kinematics (IK) and motion retargeting. The stereo camera's stream is transmitted to the VR headset as an immersive display in real-time.

\begin{figure}[t]
    \centering
    \vspace{0.1in}
\includegraphics[width=0.8\linewidth]{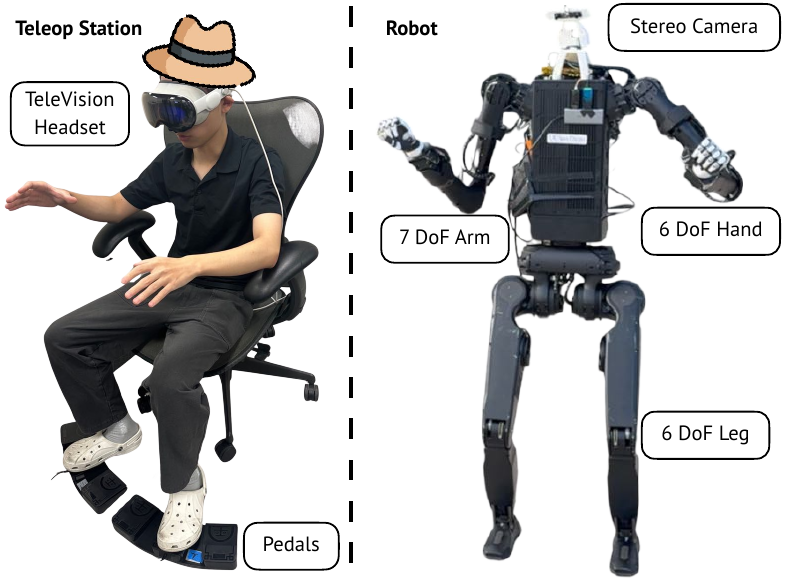}
\caption{Left: A unified teleoperation setup using Apple Vision Pro for upper body control and pedals for lower body locomotion. Right: The H1 robot with a customized head, neck, and dual 6-DoF dexterous hands, featuring an active neck and stereo camera for immersive teleoperation \cite{cheng2024tv}.}
    \label{fig:hardware}
    \vspace{-0.3in}
\end{figure}

For the lower body control, since our policy receives $\expval{\vb v,rpy,h}$ as its locomotion goal, it is compatible with any input device capable of providing these commands. Our system offers two different teleoperation setups for locomotion control. The first involves separate control: one operator controls the upper body using the VR headset, while another operator controls the lower body using a remote controller with joysticks. The second setup features a unified control, where a single user who controls the upper body also controls the lower body using pedals, as depicted in Figure \ref{fig:hardware} left. This unified setup allows the entire humanoid to be teleoperated by one person. Our teleoperation system is capable of performing a diverse set of loco-manipulation tasks, as shown in Figure \ref{fig:teaser}.

\subsection{Real-World Results}
To answer \textbf{Q3}, we conduct experiments to evaluate the locomotion robustness and manipulation precision.
\subsubsection{Robustness Test}
\begin{figure}[t]
    \centering
    \vspace{0.1in}
    \includegraphics[width=0.8\linewidth]{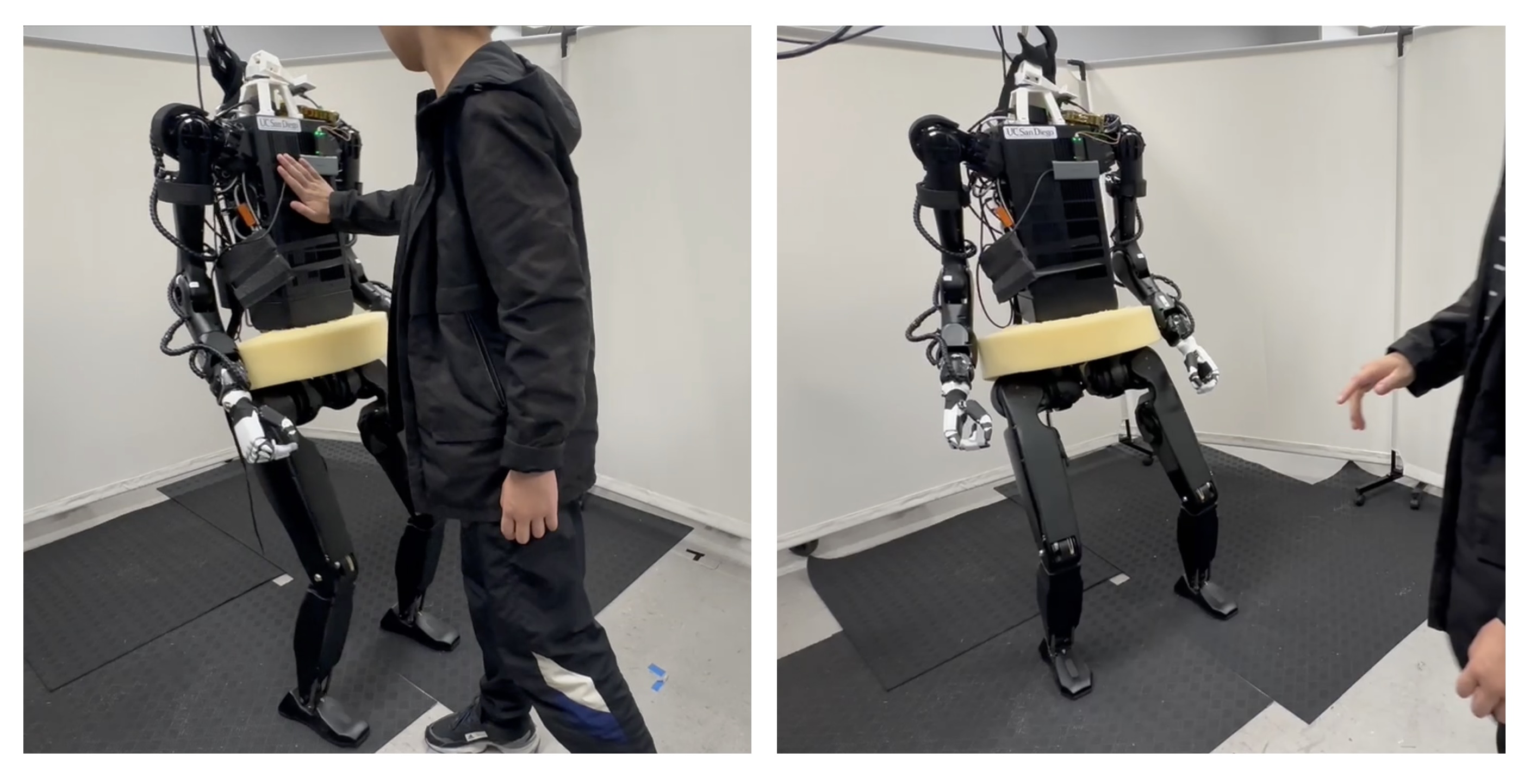}
    \caption{Robustness testing. Left: robot being pushed while standing; Right: robot recovers to the stable standing pose.}
    \label{fig:robustness}
    \vspace{-0.1in}
\end{figure}

To demonstrate that our method maintains stable locomotion control, we conduct a robustness test, as in Figure \ref{fig:robustness}. The robot's robustness is evaluated by applying a pushing force to its torso and measuring the maximum pitch velocity, the maximum pitch deviation, and the recovery time. The policies tested include \textit{Ours} (\methodn) and \textit{Ours} (\methodn) \textit{w/o motion prior}. We present the robustness testing results for both walking and standing modes separately, as shown in Table \ref{tab:robustness}. 

When pushed while walking, both policies recover equally well. When pushed while standing, the motion-prior policy shows less deviation and recovers more quickly. Standing recovery generally takes longer than walking, as the robot may need extra steps to regain stability if it initially fails to find a stationary leg position. The motion-prior policy effectively reduces both the frequency and duration of this adjustment process.

\subsubsection{Precision Evaluation}
We evaluate the precision of our method in arm tracking by measuring the distance between the tracking objectives (left \& right wrist poses tracked by the Apple Vision Pro) and the actual poses of the robot wrists in the real world. Since directly measuring the real-world poses is intractable without accurate motion capture devices, we instead use poses computed through forward kinematics based on arm motor readings. The tracking distances are reported in Table \ref{tab:precision}. The results are computed along trajectories collected while the robot performs the four tabletop tasks as described in \cite{cheng2024tv}. The results suggest that our system closely tracks the movement of human wrists. Notably, there are two layers of smoothing operations between the human wrist poses and robot joint angles: a linear filter applied to the wrist poses and PD controllers. Interactions with objects also exaggerate the tracking errors. Despite all these factors, our system consistently maintains a low tracking error.

\section{Related Work}
\label{sec:related}

\textbf{Whole-Body Control for Humanoid Robots.}  Humanoid robots need both robust mobility and precise manipulation. Traditionally, this is achieved through dynamics modeling and control~\cite{miura1984dynamic,yin2007simbicon,hutter2016anymal,moro2019whole,dariush2008whole,kajita20013d,westervelt2003hybrid}.  Recent deep reinforcement learning (RL) methods enable complex locomotion skills for legged robots~\cite{margolis2022rapid,kumar2021rma,fu2021minimizing,fu2022learning,Escontrela22arXiv_AMP_in_real,li2021reinforcement,siekmann2021blind}.  Whole-body loco-manipulation from high-dimensional sensory inputs has been explored for quadrupeds~\cite{fu2022learning,cheng2023extremeparkour,cheng2023legmanip,ha2024umi} and humanoids~\cite{cheng2024exbody,he2024h20,zhang2024wococo}. \cite{cheng2024exbody,fu2024humanplus,he2024omnih2o} train whole-body policies for robustness, employing different strategies: \cite{cheng2024exbody} decouples upper and lower body objectives to address human-robot morphological differences, \cite{fu2024humanplus} trains transformers for whole-body control and imitation learning, and \cite{he2024omnih2o} develops goal-conditioned policies. However, these studies focus on humanoids with limited arm DoFs (4–5) and do not analyze how upper-body precision affects manipulation.

\textbf{Loco-Manipulation.}
Quadrupedal robot can either use their legs for manipulation~\cite{cheng2023legmanip, lin2024locoman, ji2023dribblebot,jeon2023learning,he2024learning}, or with an additional mounted manipulator~\cite{fu2022learning, ha2024umi, liu2024visual, ma2022mani,pan2024roboduet}. Humanoid robots naturally have two arms for manipulation and two legs for locomotion. \cite{seo2023deep} uses hierarchical visual-motor policy with implicit-hierarchical whole-body control to perform fine manipulation. However, no robust walking is shown in the real world, and the focus is on static manipulation. \cite{dao2024sim} studies how to complete box loco-manipulation tasks with sim-to-real techniques and cannot generalize to various tasks.

\textbf{Motion Representation Learning.}
Human and robot motions are in high-dimensional temporal-spatial space. A good representation of motion facilitates motion understanding, generation, and imitation. Adversarial imitation methods such as \cite{peng2021amp, 2022-TOG-ASE, tessler2023calm, InterPhysHassan2023} work well with small motion datasets but suffer from mode collapse as the motions scale. 
\cite{luo2024universal, ling2020character, zhang2023vid2player3d, wang2024strategy} learn reusable motion representations via conditional variational encoders in simulation to generate novel motions. However, how CVAE can coordinate whole-body movements for real humanoid robots has not been well studied.

\section{Discussion and Limitation}
\label{sec:conclusion}
This paper presents a novel whole-body controller for humanoid robots, separately modeling the upper body with IK and retargeting, and the lower body with RL. To unify control, we introduce Predictive Motion Priors, encoding upper-body information for lower-body RL training. Our approach achieves superior performance in simulation, real-robot control, and teleoperation.

While enabling diverse loco-manipulation tasks, our method's separation of upper- and lower-body control limits agility. Additionally, humanoid hardware lacks the degrees of freedom needed for agile whole-body motion tracking. The policy handles multiple input commands, increasing the teleoperator's cognitive load. A more intuitive human-robot interaction interface is needed to mitigate this. 

\section{Acknowledgements}
This project was supported, in part, by gifts from Amazon, Qualcomm and Meta.

\bibliographystyle{IEEEtran}
\bibliography{main}

\end{document}